\documentclass{article}
\usepackage[final]{gpsmdms_2022}

\usepackage[utf8]{inputenc}

\usepackage[T1]{fontenc}    %
\usepackage[colorlinks=true]{hyperref}       %
\usepackage{url}            %
\usepackage{amsfonts}       %
\usepackage{nicefrac}       %
\usepackage{amsmath,amssymb}
\usepackage[aboveskip=2pt]{subcaption}
\usepackage{graphics}
\usepackage{xcolor}
\usepackage{mathtools}
\usepackage{bm}

\usepackage[numbers]{natbib}

\usepackage{tikz,pgfplots}
\usetikzlibrary{shapes,arrows,positioning,calc}
\usetikzlibrary{decorations.pathmorphing}
\pgfkeys{/pgf/number format/.cd,1000 sep={}}
\newlength{\figurewidth}
\newlength{\figureheight}

\usepackage{algorithm}
\usepackage{algpseudocode}

\algrenewcomment[1]{\hfill {\color{gray}\(\triangleright\) #1}}
\algnewcommand{\LineComment}[1]{\State {\color{gray}\(\triangleright\) #1}}

\usepackage[capitalize,nameinlink]{cleveref}
\crefname{section}{Sec.}{Secs.}
\crefname{appendix}{App.}{Apps.}
\crefname{algorithm}{Alg.}{Algs.}

\usepackage{xspace}
\newcommand{\eg}{\textit{e.g.}\@\xspace}
\newcommand{\ie}{\textit{i.e.}\@\xspace}

\usepackage{bm}
\newcommand{\mathbold}[1]{\bm{#1}}
\newcommand{\mbf}[1]{\mathbf{#1}}

\newcommand{\vtheta}[0]{\mathbold{\theta}}

\newcommand{\vmu}[0]{\mathbold{\mu}}
\newcommand{\vlambda}[0]{\mathbold{\lambda}}

\newcommand{\MSigma}[0]{\mathbold{\Sigma}}

\newcommand{\MZ}{\mbf{Z}}
\newcommand{\MV}{\mbf{V}}
\newcommand{\MX}{\mbf{X}}
\newcommand{\MA}{\mbf{A}}
\newcommand{\T}{\top}

\newcommand{\vz}{\mbf{z}}
\newcommand{\vf}{\mbf{f}}
\newcommand{\vm}{\mbf{m}}
\newcommand{\vu}{\mbf{u}}
\newcommand{\vx}{\mbf{x}}
\newcommand{\vy}{\mbf{y}}

\newcommand{\GP}{\mathcal{GP}}

\newcommand{\MKzz}{\mbf{K}_{\mbf{z}\mbf{z}}}
\newcommand{\MKxx}{\mbf{K}_{\mbf{x}\mbf{x}}}

\newcommand{\MKxz}{\mbf{K}_{\mbf{x}\mbf{z}}}
\newcommand{\MLambda}[0]{\mathbold{\Lambda}}
\definecolor{matplotlib-blue}{HTML}{1f77b4}

\title{Fantasizing with Dual GPs in Bayesian \\ Optimization and Active Learning}
\author{%
  Paul E.\ Chang\thanks{Work done while at Secondmind.ai}\\
  Aalto University\\
  Espoo, Finland\\
  \texttt{paul.chang@aalto.fi} \\[-1em]
  \And
  Prakhar Verma \\
  Aalto University\\
  Espoo, Finland %
  \And
  ST John \\
  Aalto University\\
  Espoo, Finland \\
  \phantom{\texttt{paul.chang@aalto.fi}} \\[-1em]
  \AND
  Victor Picheny  \\
  Secondmind.ai \\
  Cambridge, UK %
  \And
  Henry Moss  \\
  Secondmind.ai\\
  Cambridge, UK %
  \And
  Arno Solin \\
  Aalto University\\
  Espoo, Finland %
}

\begin{document}

\maketitle

\begin{abstract}
 Gaussian processes (GPs) are the main surrogate functions used for sequential modelling such as Bayesian Optimization and Active Learning. Their drawbacks are poor scaling with data and the need to run an optimization loop when using a non-Gaussian likelihood. In this paper, we focus on `fantasizing' batch acquisition functions that need the ability to condition on new fantasized data computationally efficiently. By using a sparse Dual GP parameterization, we gain linear scaling with batch size as well as one-step updates for non-Gaussian likelihoods, thus extending sparse models to greedy batch fantasizing acquisition functions. %
\end{abstract}

\section{Introduction}

Bayesian Optimization (BO, \cite{mockus1978application}) is a popular technique for black-box optimization%
.
Active Learning (AL, \cite{bect2012sequential}) is the closely related problem of learning a decision boundary. AL and BO combine Gaussian process (GP) surrogate models to predict the quantity of interest with acquisition functions that encode trade-offs between exploration and exploitation.  The maximization of these acquisition functions recommends a sequence of new points to query.

Standard GPs have a posterior distribution with a closed-form expression that can be efficiently updated \citep{chevalier2014corrected} when new observations are added to the training data. This desirable updating property has been used to design BO and AL algorithms that query $n_\textrm{new}$ points at once, by greedily building a batch of points, `fantasizing' for each point an observation value and updating the posterior \citep{ginsbourger2010kriging,desautels2014parallelizing}. 
Other BO and AL algorithms directly rely on update formulas to compute acquisition functions in closed form (\eg, \cite{chevalier2014fast,picheny2015multiobjective,wu2016parallel}).
Such formulas are not available in more complex scenarios, in particular sparse GPs \citep{titsias2009variational} or for non-conjugate likelihoods. Sparse GP models are becoming increasingly popular as surrogate models for high-throughput BO loops \cite{vakili2021scalable, moss2022information,picheny2022bayesian} where data volumes are high. 
\looseness-1

Recently, \citet{maddox2021conditioning} proposed the OVC method for the fast conditioning of sparse variational GP models, building on streaming GP approximations \cite{bui2017streaming}. They applied their OVC to evaluate existing batch acquisition functions \cite{wu2016parallel, moss2021gibbon}, showing superior performance to many existing batch BO methods. However, in order to apply their method in settings requiring non-Gaussian likelihoods, \eg~batch active learning strategies for binary classification, they must use an additional Laplace approximation with cost $O\big((m+n_\textrm{new})^3\big)$. \looseness-1

In this paper, we build on the work of \cite{maddox2021conditioning} by looking at fast conditioning of sparse GPs, but instead using the dual parameter formulation of \citet{adam2021dual}. Dual-parameterized GPs were initially motivated by a tighter hyperparameter bound and efficient natural gradient steps for optimizing the variational parameters. Here we show that they also offer decisive advantages for posterior update computations: no need for additional approximations, and the computational complexity $O(m^3+m^2 n_\textrm{new})$ remains linear in the number of conditioning points. We find that we can perform effective one-step updates for most BO and AL problems, removing the need to run a variational optimization loop with non-Gaussian likelihoods.

\begin{figure*}[t!]
  \centering
  \setlength{\figurewidth}{.27\textwidth}
  \setlength{\figureheight}{\figurewidth}
  \def\datapath{./fig/banana/probability}
  \pgfplotsset{scale only axis,axis on top}
  \begin{subfigure}[b]{.3\textwidth}
    \centering
    \begin{tikzpicture}[inner sep=0, outer sep=0]
      \node{% This file was created by tikzplotlib v0.9.8.
\begin{tikzpicture}

\begin{axis}[
height=\figureheight,
scaled x ticks=manual:{}{\pgfmathparse{#1}},
scaled y ticks=manual:{}{\pgfmathparse{#1}},
tick align=outside,
width=\figurewidth,
x grid style={white!69.0196078431373!black},
xmajorticks=false,
xmin=-2.8, xmax=2.8,
xtick style={color=black},
xticklabels={},
y grid style={white!69.0196078431373!black},
ymajorticks=false,
ymin=-2.8, ymax=2.8,
ytick style={color=black},
yticklabels={}
]
\addplot graphics [includegraphics cmd=\pgfimage,xmin=-3.80132450331126, xmax=3.6158940397351, ymin=-3.72715231788079, ymax=3.69006622516556] {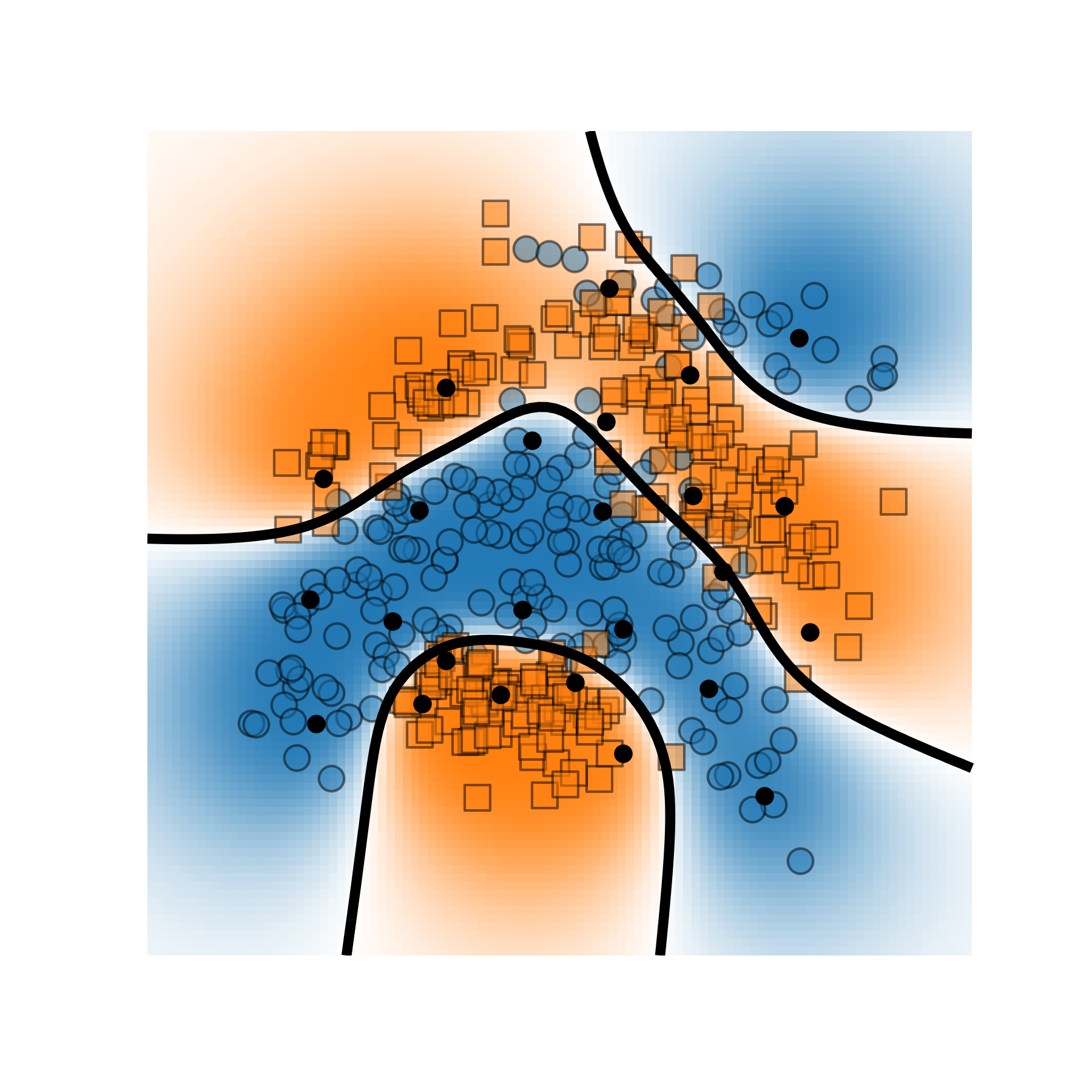};
\end{axis}

\end{tikzpicture}};
    \end{tikzpicture}
    \captionsetup{justification=centering}
    \caption{Offline SVGP model \\ (baseline)}
    \label{fig:banana_full_tsvgp}
  \end{subfigure}
  \hfill
  \begin{subfigure}[b]{.27\textwidth}
    \centering
    \setlength{\figurewidth}{.333\textwidth}
    \setlength{\figureheight}{\figurewidth}
    \begin{tikzpicture}[inner sep=0, outer sep=0]
      \foreach \x [count=\i] in {1,2,3} {
         \node (n\i) at (\i\figurewidth,0) {\input{fig/banana/probability/streaming_banana_tsvgp_\x}};
      }
      \setlength{\figurewidth}{\textwidth}
      \setlength{\figureheight}{\figurewidth}
      \node (n4) at (.666\figurewidth,-.666\figurewidth) {% This file was created by tikzplotlib v0.9.8.
\begin{tikzpicture}

\begin{axis}[
height=\figureheight,
scaled x ticks=manual:{}{\pgfmathparse{#1}},
scaled y ticks=manual:{}{\pgfmathparse{#1}},
tick align=outside,
width=\figurewidth,
x grid style={white!69.0196078431373!black},
xmajorticks=false,
xmin=-2.8, xmax=2.8,
xtick style={color=black},
xticklabels={},
y grid style={white!69.0196078431373!black},
ymajorticks=false,
ymin=-2.8, ymax=2.8,
ytick style={color=black},
yticklabels={}
]
\addplot graphics [includegraphics cmd=\pgfimage,xmin=-3.80132450331126, xmax=3.6158940397351, ymin=-3.72715231788079, ymax=3.69006622516556] {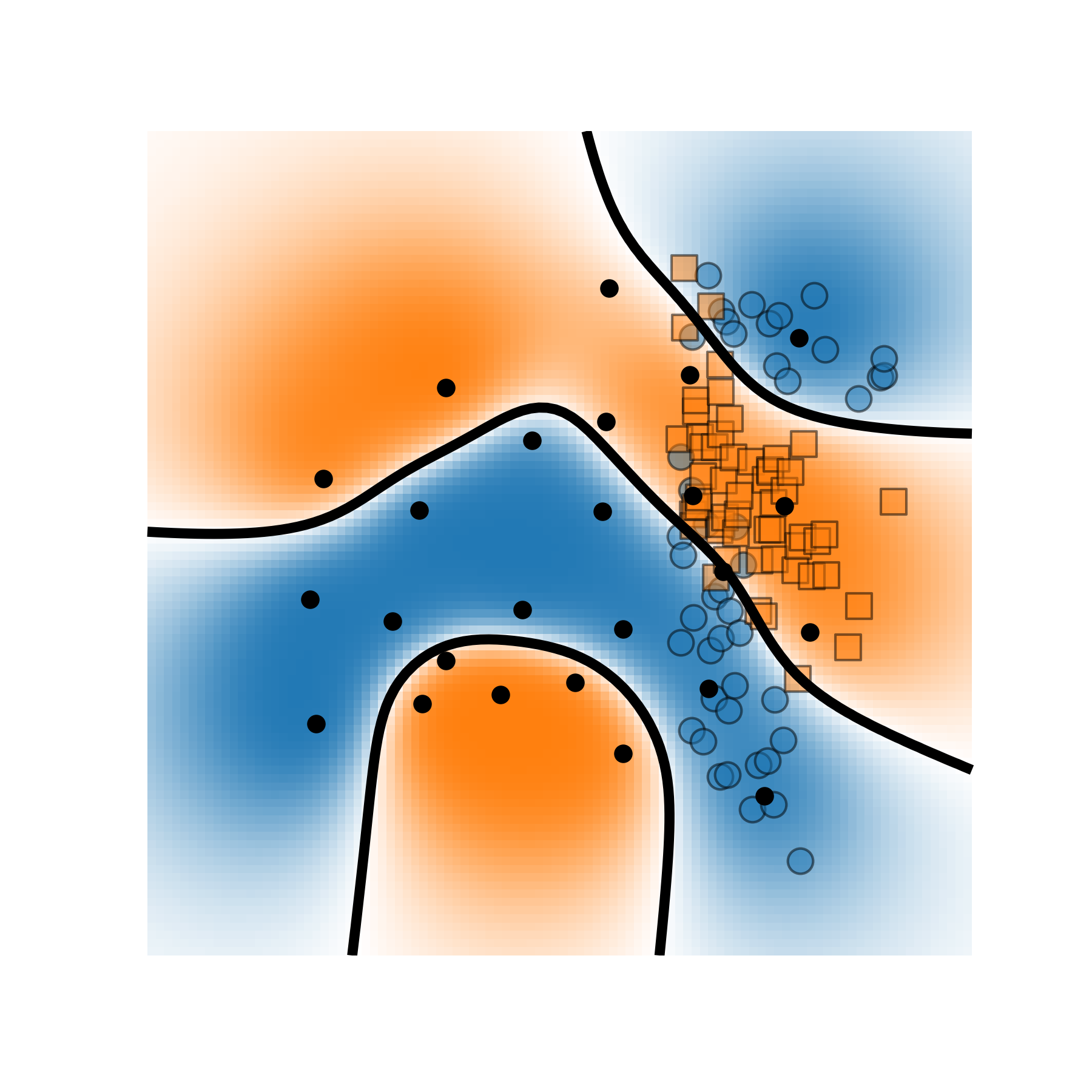};
\end{axis}

\end{tikzpicture}};
      \foreach \i in {1,2,3,4}
        \node[anchor=north west,inner sep=1pt,fill=white,draw=black] at (n\i.north west) {\i};
    \end{tikzpicture}
    \captionsetup{justification=centering}
    \caption{Streaming dual SVGP \\ (ours)}
    \label{fig:banana_online_tsvgp}
  \end{subfigure}
  \hfill
  \begin{subfigure}[b]{.27\textwidth}
    \centering
    \setlength{\figurewidth}{.333\textwidth}
    \setlength{\figureheight}{\figurewidth}
    \begin{tikzpicture}[inner sep=0, outer sep=0]
      \foreach \x [count=\i] in {1,2,3} {
         \node (n\i) at (\i\figurewidth,0) {\input{fig/banana/probability/streaming_banana_online_vargp_\x}};
      }
      \setlength{\figurewidth}{\textwidth}
      \setlength{\figureheight}{\figurewidth}
      \node (n4) at (.666\figurewidth,-.666\figurewidth) {% This file was created by tikzplotlib v0.9.8.
\begin{tikzpicture}

\begin{axis}[
height=\figureheight,
scaled x ticks=manual:{}{\pgfmathparse{#1}},
scaled y ticks=manual:{}{\pgfmathparse{#1}},
tick align=outside,
width=\figurewidth,
x grid style={white!69.0196078431373!black},
xmajorticks=false,
xmin=-2.8, xmax=2.8,
xtick style={color=black},
xticklabels={},
y grid style={white!69.0196078431373!black},
ymajorticks=false,
ymin=-2.8, ymax=2.8,
ytick style={color=black},
yticklabels={}
]
\addplot graphics [includegraphics cmd=\pgfimage,xmin=-3.80132450331126, xmax=3.6158940397351, ymin=-3.72715231788079, ymax=3.69006622516556] {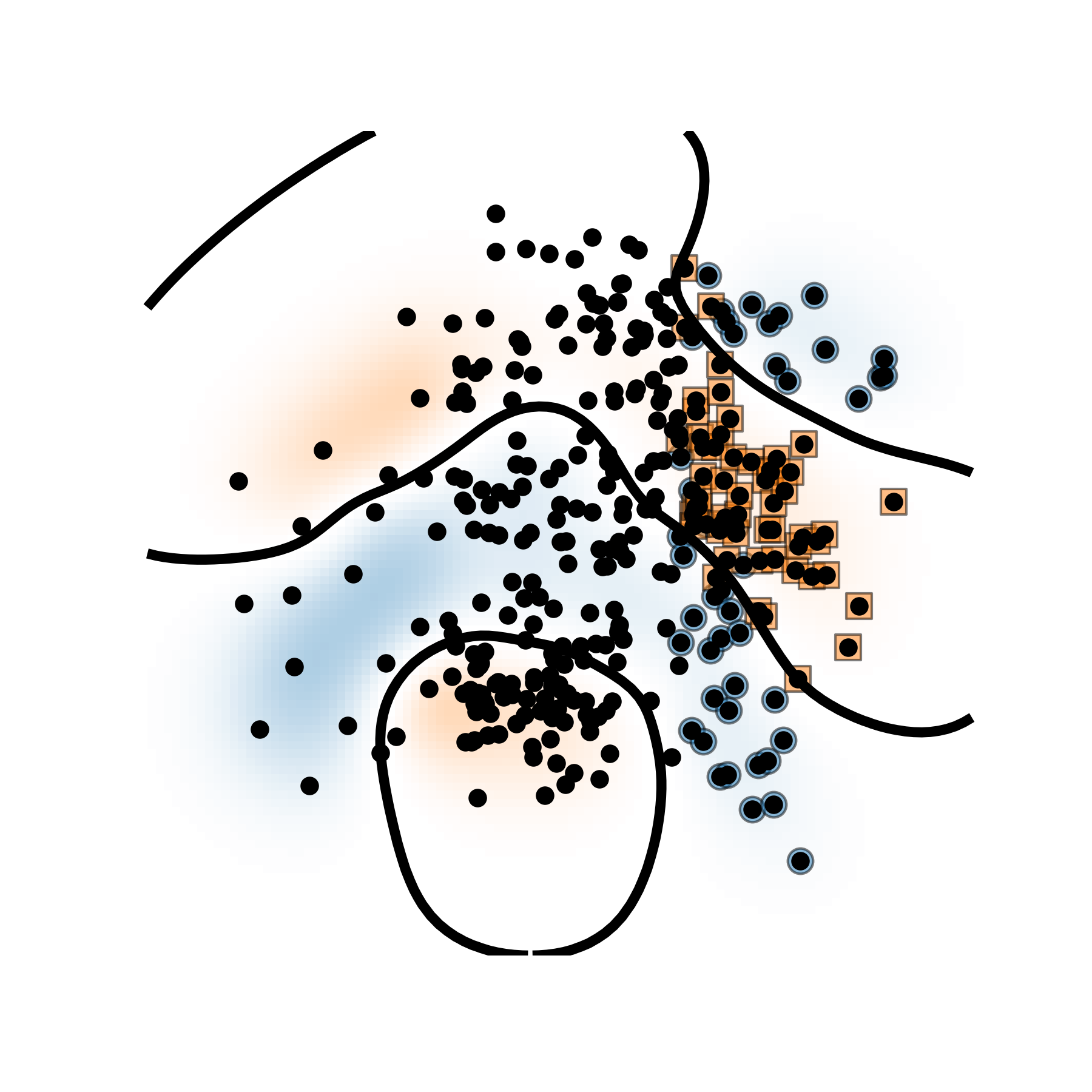};
\end{axis}

\end{tikzpicture}};
      \foreach \i in {1,2,3,4}
        \node[anchor=north west,inner sep=1pt,fill=white,draw=black] at (n\i.north west) {\i};
    \end{tikzpicture}
    \captionsetup{justification=centering}
    \caption{OVC\\ (previous work \cite{maddox2021conditioning})}
    \label{fig:banana_online_vargp}
  \end{subfigure}
  \caption[Dual conditioning on streaming banana data set]{Dual conditioning on streaming banana data set; data
  \tikz[baseline=-.6ex, opacity=0.8]{\draw[black, fill=orange] (0, 0) rectangle ++(4pt,4pt);\draw [black, fill=matplotlib-blue] (4pt, 0) circle (2pt);}
  appears batch by batch (1--4).
  The plot shows the decision boundary
  \tikz[baseline=-.6ex,line width=1.5pt]\draw[black](0,0)--(.5,0);
  and the predictive class probability, with colour shading
  \tikz[baseline=1pt]\node[rectangle, anchor=south, right color=orange, left color=white, minimum width=12pt, minimum height=6pt]{};
  and
  \tikz[baseline=1pt]\node[rectangle, anchor=south, right color=matplotlib-blue, left color=white, minimum width=12pt, minimum height=6pt]{};
  increasing the more certain the model is about the class. The inducing points are overlaid as black dots. (a)~Dual SVGP model trained with full data. (b)~Dual SVGP model conditioned on the data appearing in batches. (c)~ Online Variational Conditioning (OVC, \cite{maddox2021conditioning}) model on batched data.\looseness-1} 
  \label{fig:banana_conditioning_comparison}
\end{figure*}

\section{Methods}
\label{sec:methods}
Gaussian process models define a prior over (latent) functions $f \sim \GP(0, \kappa)$ that is characterized completely by a covariance function $\kappa(\vx,\vx')$. When a GP prior is combined with a data set ${\mathcal{D} = (\MX,\vy) =\{(\vx_i, y_i)\}_{i=1}^n}$ of input--output pairs and a Gaussian likelihood, computational complexity of the posterior process is $O(n^3)$. If the likelihood is non-conjugate, a common technique is to perform variational inference. Fitting a Variational GP model (VGP, \eg \cite{khan2017conjugate}) requires multiple $O(n^3)$ operations in an optimization loop until convergence.
Sparse Variational GP (SVGP) models \citep{titsias2009variational, hensman2015scalable} overcome the cubic scaling while simultaneously dealing with non-Gaussian likelihoods. They swap computations on the full training set $\MX$ with a sparser set of \textit{inducing points} $\MZ \coloneqq (\vz_j)_{j=1}^m$, $(\vu)_j = f(\vz_j)$, with the approximate posterior $q(\vu ; \vm,\MV)$. The overall complexity is $O(nm^2)$, where $m \ll n$. The induced posterior of the function values at all other points is:
\begin{equation}
    q_{\vu}(\vf) = \mathrm{N}(\vf ; \MA\vm^*, \MKxx - \MA \MKzz^{-1} \MA^{\T} + \MA \MV^* \MA^{\T}), \label{eq:prediction}
\end{equation}
where $\MKxx$ is an $n \times n$ matrix with $\kappa(\vx_i,\vx_j)$ as the $ij$\textsuperscript{th} entry,  $\MA = \MKxz \MKzz^{-1}$. $\MKxz$ and $\MKzz$ are defined similarly to $\MKxx$. To predict with an SVGP we need to infer the variational parameters $(\vm^*,\MV^*)$ at the optimum of the evidence lower bound \citep[ELBO,][]{titsias2009variational}. The ELBO is also used to learn the hyperparameters $\vtheta$ of kernel and likelihood. 

\citet{adam2021dual} showed that the optimal variational parameters $(\vm^*,\MV^*)$ are: 
\begin{align}
    \vm^* \equiv  \MV^* \vlambda^* \qquad \text{and} \qquad %
    \MV^* \equiv [\MKzz^{-1} + \MLambda^*]^{-1}. \label{eq:optimality_cond}
\end{align}
The optimal dual variables $(\vlambda^*,\MLambda^*)$ are found using the following iteration until convergence:
\begin{equation}\label{eq:updates_iter}
    \begin{split}
    \vlambda_{t} &= (1 -\rho_t)\vlambda_{t-1} + \rho_{t}\nabla_{\vmu^{(1)}} \mathbb{E}_{q_{\vu}(\vf)} [ \log p(\vy \mid \vf) ], \\
    \MLambda_{t} &= (1 -\rho_t)\MLambda_{t-1} + \rho_{t}\nabla_{\vmu^{(2)}} \mathbb{E}_{q_{\vu}(\vf)} [ \log p(\vy \mid \vf) ], 
    \end{split}
\end{equation}
where $(\vmu^{(1)},\vmu^{(2)})$ are the expectation parameters of the approximating distribution $q(\vu)$, so $\vmu^{(1)}=\vm$ and $\vmu^{(2)}= \MV +\vm\vm^\T$, and $\vlambda_{0}$ and $\MLambda_{0}$ are initialized to zero. As pointed out in \cite{khan2017conjugate}, these are in fact natural gradient updates, which \cite{adam2021dual} and \cite{chang2020fast} showed to be computationally faster than existing natural gradient methods and to have better convergence than the $(\vm,\MV)$ parameterization.
\begin{algorithm}[t!]
  \caption{Fantasizing a batch with Dual Conditioning.} \label{alg:BO/AL}
  \textbf{Input:} $(\vlambda^*, \MLambda^*)$, $k$, $\alpha(\cdot)$ \\
  \textbf{Initialize:} $\MX_b = \emptyset$
    \begin{algorithmic}[1] 
    \For{$i$ in $1,2,\ldots,k$}\Comment{$k$ is desired number of query points}
    \State $\vx_i = \mathrm{arg}\,\max_{\vx} \alpha(\vx)$ \Comment{Calculate $\alpha(\vx)$ using prediction function \cref{eq:prediction} at $\vx$}
    \State $y_i = \mathbb{E}[f(\vx_i)]$ \Comment{Fantasized $y$ is mean of the GP at $\vx_i$}
    \State $\mathcal{D}_{\textrm{new}} = (\vx_i,y_i)$ \Comment{The fantasized data point is treated as new data}
    \State Compute $(\vlambda^*, \MLambda^*)$ using \cref{eq:fast_condit} and $\mathcal{D}_{\textrm{new}}$ \Comment{Dual conditioning}
    \State $\MX_{\textrm{b}} \leftarrow \MX_{\textrm{b}} \cup \vx_i$ \Comment{$\vx_i$ is added to the current batch points}
    \EndFor
  \end{algorithmic}
  \textbf{Return:} $\MX_{\textrm{b}}$ \Comment{$\MX_{\textrm{b}}$ is the chosen batch of points.}
\end{algorithm}
In the conditioning problem, the starting point is a set of trained SVGP variational parameters $(\vm_{\textrm{old}},\MV_{\textrm{old}})$ found using ${\mathcal{D}_{\textrm{old}} = (\MX_{\textrm{old}},\vy_{\textrm{new}})}$ of size $n_{\textrm{old}}$. As new data $\mathcal{D}_{\textrm{new}} = (\MX_{\textrm{new}},\vy_{\textrm{new}})$ of size $n_{\textrm{new}}$ is available, the variational parameters are updated to reflect $\mathcal{D}_{\textrm{old}} \cup \mathcal{D}_{\textrm{new}}$. Furthermore, in the conditioning problem, $\vtheta, \MZ$ are assumed to be constant. In a na\"{i}ve implementation, specifically using an $(\vm,\MV)$ parameterization, it is not obvious how to solve the conditioning problem. 

In dual parameterization, the problem of conditioning on new data becomes easier. We use \cref{eq:updates_iter} on just the new data to find $(\vlambda^*_{\textrm{new}},\MLambda^*_{\textrm{new}})$. In conjugate models with Gaussian likelihoods, we can simply add this to the old $(\vlambda^*_{\textrm{old}},\MLambda^*_{\textrm{old}})$, saving the need to recompute the dual parameters at old data. Thus \cref{eq:updates_iter} is a one-step update (\ie, $\rho=1, t=1$). We call this method \emph{Dual Conditioning}. In sparse Gaussian process regression (SGPR), the change in computation of the posterior process in \cref{eq:prediction} is from $O((n_\textrm{new}+n_\textrm{old})m^2 + )$ to $O(n_\textrm{new}m^2)$, which can be a substantial saving when needing to fantasize and condition points in batch acquisition functions. The SGPR formulation of Dual Conditioning was first presented in \cite{bui2017streaming} using the idea of pseudo-data $(\tilde{\vy},\tilde{\MSigma})$ which is the source parameter form of $(\vlambda^*_{\textrm{old}},\MLambda^*_{\textrm{old}})$. \citet{maddox2021conditioning} show the usefulness of this computational saving for batch BO.

In the non-conjugate setting, \cite{maddox2021conditioning} resorts to a Laplace approximation to find the dual parameters, which must be run for multiple iterations with a complexity\footnote{See the `Application to BO' section in \citet{maddox2021conditioning}.} of $O\big((m+n_\textrm{new})^3\big)$. %
Our non-Gaussian Dual Conditioning method has a superior computational complexity of $O(n_\textrm{new}m^2 +m^3)$ per iteration and uses natural gradient variational inference, which outperforms the Laplace approximation \cite{khan2021bayesian}. In this setting, our previous posterior $q(\vu; \mathcal{D}_{old})$ already contains the bulk of data, thus allowing the one-step updating of $(\vlambda^*_{\textrm{new}},\MLambda^*_{\textrm{new}})$ even in the non-Gaussian setting ($t=1$ and choosing $\rho=1$ in \cref{eq:updates_iter}). This update step is given by Dual Conditioning updates:\looseness-1
\begin{equation} \label{eq:fast_condit}
\begin{split}
    \vlambda^* \leftarrow & \vlambda^*_{\textrm{old}} + \nabla_{\vmu^{(1)}} \mathbb{E}_{q_{\vu}(\vf_{\textrm{new}})} [ \log p(\vy_{\textrm{new}} \mid \vf_{\textrm{new}})],  \\
        \MLambda^* \leftarrow & \MLambda^*_{\textrm{old}} + \nabla_{\vmu^{(2)}} \mathbb{E}_{q_{\vu}(\vf_{\textrm{new}})} [ \log p(\vy_{\textrm{new}} \mid \vf_{\textrm{new}})]. 
\end{split}
\end{equation}
\cref{fig:banana_conditioning_comparison} shows that using this form of Dual Conditioning outlined in \cref{eq:fast_condit} we are able to recover a very similar posterior to the offline solution. To %
build a batch for BO and AL applications, we follow the Kriging Believer algorithm (see Alg.~1 on p.~17 in \cite{chevalier2014fast}): we fantasize data points given from an acquisition function $\alpha(\cdot)$ and then use \cref{eq:fast_condit} to add the fantasized points. We outline our algorithm in \cref{alg:BO/AL} and optimize $\alpha(\cdot)$ using \cite{wilson2018maximizing}. We keep inducing inputs $\MZ$ and hyperparameters $\vtheta$ fixed within this inner loop of optimizing the acquisition function, but update them in the outer loop when we obtain new experimental data.

\section{Experiments}
To compare the fast conditioning of the dual SVGP model with that of the OVC model \cite{maddox2021conditioning}, we investigate a streaming binary classification task. We also demonstrate the benefits of our approach on a high-dimensional stochastic Bayesian optimization problem.

{\bf The streaming banana classification experiment} is the same as in \cite{maddox2021conditioning}. The data set is divided into four batches of $100$ points each. In each step, only the current batch and previously inferred variational parameters are accessible; %
therefore, one must use an online model to condition on new data. %
We compare the decision boundary and predictive class probability for three models in \cref{fig:banana_conditioning_comparison}. As an oracle baseline, an offline SVGP model with $25$ inducing points and Mat\'ern-$\nicefrac{5}{2}$ kernel is trained on the full data, see \cref{fig:banana_full_tsvgp}. Our method uses Dual Conditioning (\cref{eq:fast_condit}) on the streaming data (batches 1--4). For the OVC method, we run the \href{https://github.com/wjmaddox/online_vargp/blob/main/notebooks/streaming_bananas_plots.ipynb}{code published by \citet{maddox2021conditioning}}. Their method essentially initializes new models on each batch and then combines them, hence the increasing number of inducing points. The evolution of the class probability of dual SVGP and OVC is shown in \cref{fig:banana_online_tsvgp} and \cref{fig:banana_online_vargp}. The class probability obtained by the dual SVGP model after seeing the final batch closely matches that of the offline SVGP model. In contrast, the OVC method does not recover the full-data decision boundary, and its uncertainty does not match the offline baseline well.\looseness-1

\begin{figure*}[t!]
  \centering\footnotesize
  \def\datapath{./fig/lunar/}
  \centering
  \setlength{\figurewidth}{.4\textwidth}
  \setlength{\figureheight}{.8\figurewidth}
  \pgfplotsset{scale only axis,axis on top, yticklabel style={rotate=90,anchor=base,yshift=0.2cm}}
  \vspace*{-1em}
  \begin{subfigure}[b]{.48\textwidth}
    \centering
    \scriptsize
    \input{fig/lunar/lunar_lander_reward_observations}
  \end{subfigure}
  \hfill
  \begin{subfigure}[b]{.48\textwidth}
    \centering\scriptsize
    \begin{tikzpicture}[inner sep=0]
      \node[anchor=north west] at (0,0) {\includegraphics[width=6cm,trim=0 100 0 0,clip]{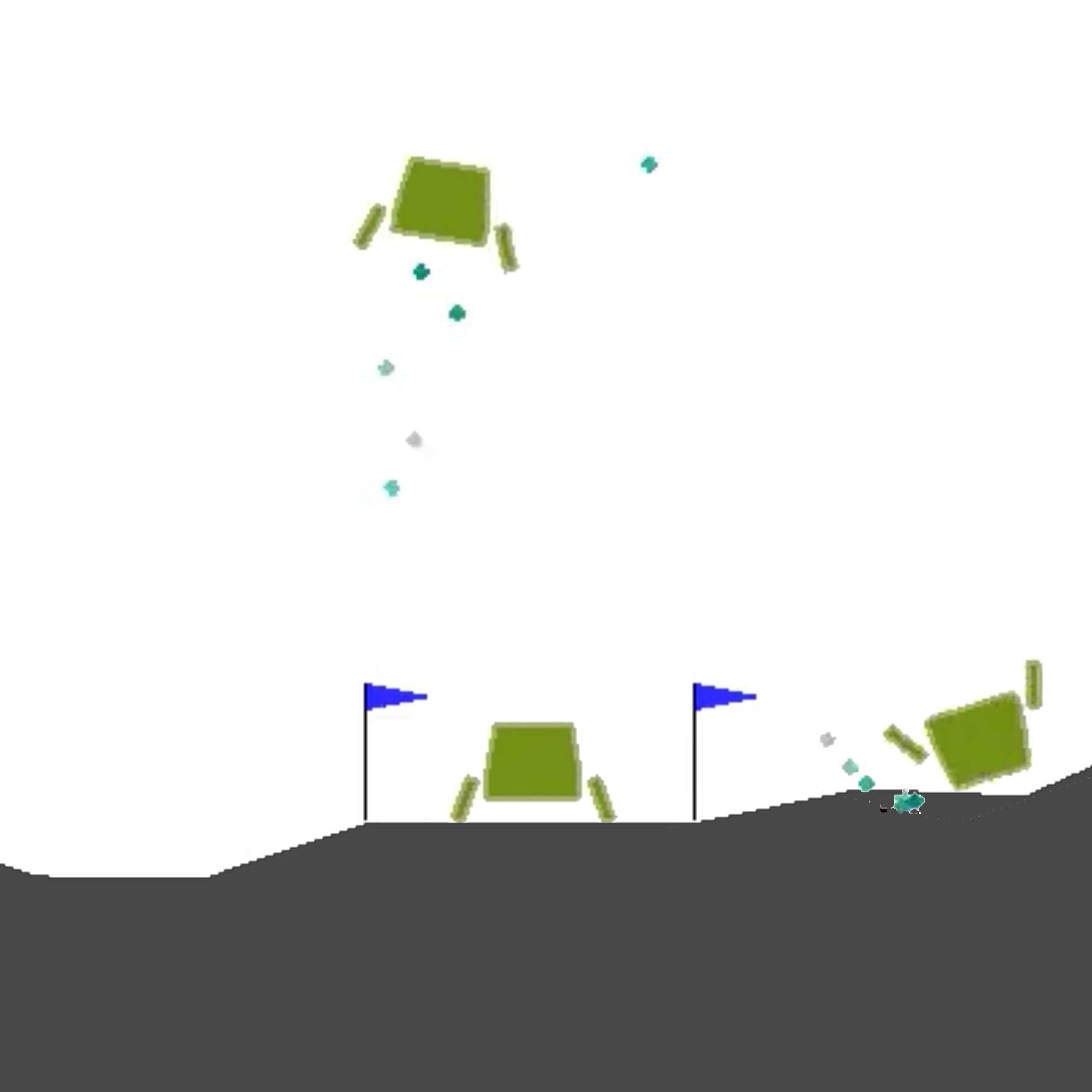}};

      \coordinate (a) at (2.45,-1.2);

      \draw[black,-latex',->,thick] (a) -- ++(0,-1);
      \draw[black,-latex',->,thick] (a) -- ++(1,0);

      \node[align=left,font=\scriptsize] at ($(a) + (2.5,-1)$) {
        State variables: \\
        $\bullet$~position: $x,y$ \\
        $\bullet$~velocity: $v_x, v_y$ \\
        $\bullet$~orientation: $\omega, \dot{\omega}$ \\
        $\bullet$~indicator variables\\
        $\bullet$~thruster states};
        
      \node at ($(a) + (0,.5)$) {Random initial state};
      \node at ($(a) + (.5,-3.7)$) {\color{white}Success};
      \node at ($(a) + (2.7,-3.5)$) {\color{white}Failure};
      \node[rotate=90] at ($(a) + (-2,-1.5)$) {Stochasticity from wind};
      \draw[->,blue!50,decoration={snake}, decorate] ($(a) + (-1.7,-1)$) -> ++(1,0);
      \draw[->,blue!50,decoration={snake}, decorate] ($(a) + (-1.7,-2)$) -> ++(1,0);
      
    \end{tikzpicture}\\[.8cm]~
    \captionsetup{justification=centering}
  \end{subfigure}
  \caption{Bayesian optimization on a lunar landing setup; the aim is to successfully land on the surface between the flags. The stochasticity in the initial state and in the environment caused due to wind makes the problem challenging. We compare the batch solution with fantasization against the non-batch solution showcasing the benefit of batch acquisition in such noisy setups.}
  \label{fig:lunar_landing}
\end{figure*}

{\bf The lunar lander problem} is a challenging rocket optimization problem that aims to land successfully in a specified target region, as considered previously by \cite{moss2020bosh,paleyes2022penalisation}. Here, every action performed by the lander results in a reward, and the aim is to optimize the total reward. Various environmental components add stochasticity, making it a challenging problem. The setup with the sources of stochasticity and multiple states of the lander is shown in \cref{fig:lunar_landing}. The search space spanning over $\mathbb{R}^{12}$ is high-dimensional; however, batching and fantasizing the data points can help overcome the difficulty. We model the problem using a regression model that aims to maximize the reward and a classification model for whether the landing was successful, using \cref{alg:BO/AL}. The acquisition function is a product of Expected Improvement of the regression model and the predictive mean of the classification model. %
The non-Gaussian likelihood of the classification model prevents the use of many batch acquisition functions which exploit properties unique to Gaussian likelihoods, such as q-KG \cite{wu2016parallel}. Our method is agnostic to the likelihood, and so allows batching and fantasizing in this challenging setup. The same set of $24$ initial data points is used and optimized for $50$ iterations for both models. The batch model builds batches of $20$ query points. We run the experiment with $10$ random initial observations and plot mean and individual rewards along with the BO iterations in \cref{fig:lunar_landing}.\looseness-1

\section{Discussion and Conclusion}
In this paper, we have unified previous work on efficiently conditioning sparse GPs on new data; our framework allows for the same variational method irrespective of the likelihood. In big-data regimes our method can be used to efficiently incorporate new information into the posterior. We show the usefulness of our technique on batch BO/AL with an experiment on a complex stochastic problem.  In future work, we wish to apply our method in model-based reinforcement learning as well as extending it to updating $\MZ$ and $\vtheta$ in the fully online setting discussed by \citep{bui2017streaming}.

\end{document}